\documentclass[acmlarge]{acmart}
\AtBeginDocument{%
  }

\usepackage{subcaption}
\usepackage{comment}
\usepackage{booktabs}
\usepackage{amsmath}
\usepackage{lineno}
\usepackage{enumitem}
\usepackage{makecell}
\usepackage{caption}
\usepackage[dvipsnames]{xcolor}
\usepackage{hyperref}
\usepackage{xurl} 
\usepackage{tablefootnote}

\setcopyright{acmlicensed}
\copyrightyear{2026}
\acmYear{2026}
\acmDOI{XXXXXXX.XXXXXXX}
\begin{document}

\title[Generative AI and the Crisis of Higher Education Assessments]{Generative AI and the Crisis of Higher Education Assessments: Examining ``AI-Proofing'' Strategies for College-Level Assessments}


\author{Bonam Mingole}
\email{bjm6940@psu.edu}
\author{Davis Yadav}
\email{kzl5736@psu.edu}
\author{Hui Min Lee}
\email{hvl5699@psu.edu}
\affiliation{%
 \institution{Pennsylvania State University}
 \city{University Park}
 \state{PA}
 \country{USA}}
 
\author{S. Shyam Sundar}
\email{sss12@psu.edu}
\affiliation{%
 \institution{Pennsylvania State University}
 \city{University Park}
 \state{PA}
 \country{USA}}
\affiliation{%
 \institution{Sungkyunkwan University}
 \city{Jongno-gu}
 \state{Seoul}
 \country{Republic of Korea}}

 \author{Amulya Yadav}
\email{amulya@psu.edu}
\affiliation{%
 \institution{Pennsylvania State University}
 \city{University Park}
 \state{PA}
 \country{USA}}

\renewcommand{\shortauthors}{Author One et al.}

\begin{abstract}
Higher education is confronting an academic-integrity challenge unlike any in its history. Students can now outsource homework, take-home assignments, and online examinations to large language models (LLMs) that complete them in seconds. Existing research on evaluating LLM performance on academic assessments has focused largely on standardized examinations, or on questions from college-level courses drawn from a single discipline. Far less is known about how LLMs perform on college-level assessment questions spanning multiple disciplines, that instructors have purposely designed to be AI-resistant, and about which features of such questions, if any, meaningfully constrain LLM performance. We address this gap by collecting 75 questions from authentic college coursework assessments, spanning 17 disciplines, which were contributed by 37 faculty members from ten U.S. universities and one U.K. university. Of these, 61 questions were intentionally designed by faculty members to be ``LLM-proof'', while the remaining 14 questions were sourced from online courses. We then (i) evaluate three state-of-the-art LLMs' ability to correctly answer these questions with and without supplementary course materials; (ii) identify five strategies, through thematic analysis, that faculty used to make their questions harder for LLMs to solve; and (iii) test whether these strategies impact LLMs' ability to answer the questions accurately. Our experimental results show that the best-performing model scored approximately 83\% without supplementary course materials and 85\% with them, and we observed noteworthy differences across models, answer formats, and instructor tweaking strategies. We discuss implications for assessment design, academic-integrity policy, and the broader ethics of deploying LLMs in educational settings.

\end{abstract}

\ccsdesc[500]{Computing methodologies~Artificial intelligence}
\ccsdesc[500]{Applied computing~Education}

\keywords{Large Language Models, AI in Education, Higher Education, AI-resistant Assessments, Academic Integrity}


\maketitle

\section{Introduction}
Since the public release of ChatGPT in late 2022 \cite{openai2022introducing}, large language models (LLMs) have become deeply embedded in students' academic workflows. Within five months of launch, surveys indicated that over 50\% of U.S. college students had used the chatbot to complete their assignments \cite{welding2023half}, and its adoption has continued to accelerate over time. This rapid uptake is further enabled by the consistent ability of AI-generated work to evade both human scrutiny and automated detection \cite{perkins2023academic}, effectively driving the marginal cost of academic misconduct towards zero, and blurring the lines between unauthorized use and legitimate assistance \cite{ravvselj2025higher}. 

In fact, students now routinely rely on LLMs across the educational lifecycle, from writing the personal statements that secure college admissions \cite{schaffer2025kaplan}, to completing the assignments and take-home exams that ultimately confer their degree \cite{donnell2024role, walsh2025everyone}. The normalization of AI assistance at these critical stages of academic training raises universal concerns for academic integrity \cite{college2026new, sullivan2023chatgpt}, a cornerstone of higher education \cite{nartgun2024cheating}. These concerns are increasingly translating into concrete institutional action. For example, higher-education institutions in the United Kingdom formally sanctioned about 7,000 students for AI-assisted cheating in the 2023--24 academic year, roughly triple the previous year's figure \cite{goodier2025revealed}, and at least four institutions expelled students for AI-related cheating in 2024--25 \cite{hogan2026surge}. More recently, the Provost of Dartmouth College had to resign following accusations that he failed to disclose AI use in his academic writing \cite{apnews2026darmouth}.

At the same time, higher-education instructors are adapting their course assessments in myriad ways to mitigate AI-assisted cheating in their courses, e.g., introducing proctored in-person exams \cite{hindy2026students}, oral examinations \cite{gecker2026perfect}, and/or modifying (or tweaking) existing assignment/exam questions from their courses to make them ``LLM-proof.''
Given the centrality of assessment to universities' educational mission, evaluating the effectiveness of these ``LLM-proofing'' strategies is essential. However, existing empirical studies largely assess LLMs using standardized examinations \cite{kung2023performance, katz2024gpt} or multiple-choice benchmarks such as MMMU \cite{yue2024mmmu}, whose questions mostly predate LLMs. Studies using authentic instructor-written questions typically do not establish whether these questions were designed to resist AI and are often restricted to narrow response formats like textual open responses \cite{borges2024could}, or focus on a single discipline \cite{stoyanov2025identifying}. Consequently, evidence remains limited on LLM performance across diverse disciplines and response formats on assessments intentionally designed to resist AI.  

To address this gap, we investigate state-of-the-art LLMs' performance on heterogeneous higher-education assessment questions which were intentionally designed by faculty members (across institutions) to be ``LLM-proof''. Our work is guided by the following research questions. \textbf{RQ1}-How effective are state-of-the-art LLMs, when used to answer college-level assessment questions that are intentionally designed to be ``LLM-proof''? \textbf{RQ2}-What types of strategies are used by instructors to create LLM-proof assessments, and how effective are such strategies in constraining LLM performance? \textbf{RQ3}-What characteristics of assessment questions, if any, affect the ability of LLMs to answer them?
To answer these questions, we conducted a crowdsourced data collection effort involving faculty participants from ten universities in the United States and one university in the United Kingdom. The resulting dataset contains 75 questions across 17 disciplines and a great diversity in question structures and answer formats. Out of these 75 questions, 61 were intentionally designed by instructors to increase their difficulty for LLMs, while the remaining 14 are sourced from online courses. Importantly, the dataset also includes 11 pairs of original questions and instructor-modified (``tweaked'') variants, enabling direct comparisons between conventional and AI-resistant question designs. Using this dataset, we first evaluate three state-of-the-art LLMs' ability to answer our crowdsourced instructor-crafted ``LLM-proof'' college-level questions, with and without supplementary course materials, using LLM-as-a-judge for scoring \cite{li2025generation}. Second, we conduct thematic analysis to uncover five strategies used by faculty to make their assessment questions LLM-proof, and analyze whether these strategies are effective in constraining LLM performance on their questions. Third, we systematically examine how question characteristics influence model performance.  

Our findings show that state-of-the-art LLMs achieve consistently high scores across our dataset of ``LLM-proof'' instructor-crafted assessments. GPT-5.4 performs best, averaging $\sim$83\% without and $\sim$85\% with supplementary course materials.  We also find that some question-tweaking strategies are more effective than others at reducing LLM performance, while answer format characteristics have limited impact. This work contributes to ongoing discussions regarding the future of higher education assessment, academic integrity, and responsible AI use in education.

\section{Related Work}
\label{sec:related_work}

\textbf{Evaluating LLMs on Academic Questions.}
Research evaluating LLMs on college-level assessments has expanded rapidly, with studies generally reporting average to high performance. Much of this work focuses on standardized examinations, including national college entrance exams \cite{liu2025matheval, park2025evaluating}, graduate admission tests \cite{zhong2024agieval}, medical question banks \cite{beaulieu2024evaluating}, and finance credentialing tests \cite{shetty2025advanced}. Other studies use large benchmarks such as MMLU \cite{wang2024mmlu}, which primarily draws questions from online datasets. Thus, these studies do not provide evidence of LLMs' ability to answer the kind of assessments students encounter in their own college courses. Recent research has addressed this limitation by evaluating LLMs on authentic college assessments. However, most evaluations focus on individual disciplines or courses, including computer science \cite{dinh2024sciex}, law \cite{choi2021chatgpt}, physics \cite{yeadon2024comparison}, and medicine \cite{stribling2024model, yoshioka2025performance}. Humanity's Last Exam (HLE) covers academic questions across dozens of disciplines \cite{center2026benchmark}, yet limits responses to exact-match strings and multiple-choice selections. Similarly, C-Eval covers 25 college-level subjects using exclusively multiple-choice items \cite{huang2023c}, while Borges et al. evaluate GPT-4 across seven disciplines using textual open-response and multiple-choice formats \cite{borges2024could}. These restrictions limit the extent to which such evaluations capture the diversity of real-world college assessments, which vary in both question structure and expected response format.

In response to the aforementioned studies, college instructors are increasingly modifying assessments to make them more resistant to AI-assisted cheating. Our study evaluates the effectiveness of these modifications using 75 authentic college assessment questions spanning 17 disciplines, contributed by 37 faculty members from 11 research-intensive universities (10 in the United States and one in the United Kingdom). The participating institutions include both public and private universities, and the represented courses comprise 34\% graduate-level and 66\% undergraduate-level coursework. Unlike prior benchmarks with restricted response formats, our dataset encompasses four answer formats, ranging from written expression to graphical representation, enabling a broader evaluation of LLM performance on authentic college-level assessments designed to resist AI.

\textbf{Evaluating AI-resistant Academic Questions.}
A growing body of research examines how assessments can be designed to resist AI-generated responses. Awadallah and Khlaif propose question-design recommendations without any empirical validation \cite{awadallah2024ai}, while Larsen proposes a few simple methods to adapt questions, but evaluates these methods on online benchmarks only \cite{larsen2023creating}. A few recent studies also focus on adapting instructor-authored questions, but remain confined to computer-science related disciplines \cite{akbar2025beyond, ding2025designing, elsayed2023towards}. Thus, all prior studies either focus on post-hoc adaptation of existing academic assessments, or are limited in their diversity to only a single academic discipline.


To our knowledge, our study is the first to empirically evaluate instructor-authored questions that are intentionally designed (from scratch) to resist LLM-generated responses across multiple academic disciplines. Of the 75 questions in our dataset, 61 were intentionally designed by instructors to be ``LLM-proof,'' while the remaining 14 were sourced from online courses and serve as a control group because their intended resistance to LLMs is unknown. Further, we investigate the effectiveness of specific AI-resistant question-design strategies by comparing LLM performance on original and instructor-tweaked question variants across seven academic disciplines and diverse answer formats. We also compare model performance with and without supplementary faculty-provided course materials. Together, our analyses extend prior work by evaluating both the effectiveness of AI-resistant question design and the factors that may influence LLM performance on college assessments intentionally designed to be AI-resistant.

\section{Building a Dataset of ``LLM-proof'' Questions}
To address the aforementioned research questions (\textbf{RQs} 1, 2, and 3), we organized and hosted an IRB-approved two-phase competition in Spring 2025 at a leading U.S. public research university. 
The first phase of the competition was open to faculty around the world, who were invited to submit self-authored college-level assessment questions that they had either designed from scratch or tweaked an existing assessment question to make it more ``LLM-proof''. The second phase of the competition was open to undergraduate and graduate students at the organizing university, who were invited to attempt to answer these  questions using only LLMs. The month-long event engaged participants in exploring LLMs' effectiveness in answering ``LLM-proof'' college-level assessments. Participation in the competition was encouraged through targeted outreach and promotion, e.g., repeated emails, flyers, social media posts, etc. A total of 37 faculty members from ten U.S. universities and one U.K. university participated in the first phase. In the second phase, 31 students from 10 different academic majors participated. Both the faculty and the student participants joined the competition voluntarily due to their strong interest in AI and its impact on higher education. During the informed consent process, student participants were cautioned that LLM-generated responses (to questions collected in Phase 1) could be misleading and therefore should not be regarded as reliable. All participants consented to the use of their anonymized submissions for research purposes.

\textbf{Phase 1: } During the first phase of the IRB-approved competition, faculty participants around the world were invited to submit two distinct types of self-authored questions: (i) self-authored questions designed from scratch to make them LLM-resistant, and (ii) modified questions which corresponded to tweaked versions of existing questions, to make them more LLM-resistant. In addition to submitting the question as a PDF (via an online submission form), instructors were asked to submit a mandatory solution file, and an optional grading rubric. 
Further, for questions that were tweaked versions of existing questions, instructors were asked to submit both the original and the new tweaked question. For each question that they submitted, instructors were also asked to provide a written explanation describing their rationale for how the question was designed or tweaked, and why they believe the question to be ``LLM proof''. The submission form provided to faculty participants is described in Appendix \ref{ax:faculty_submission_form}.

A total of 66 questions were submitted by 37 faculty participants (from ten U.S. universities and one U.K. university) spanning 17 academic disciplines. Faculty participants' demographic information is reported in Appendix \ref{ax:faculty_demographics}. Five submissions were excluded because they were duplicates or lacked solution files, resulting in a final corpus of 61 faculty-contributed ``LLM-proof'' questions of which 50 were questions designed from scratch, and 11 were tweaked variants of existing assessment questions.

\noindent
\begin{minipage}[t]{0.59\columnwidth}
\vspace{0pt}
\noindent 
Out of the 61 faculty-submitted ``LLM-proof'' questions, 23 (38\%) included instructor-provided grading rubrics. The research team developed rubrics for the remaining 38 (62\%) based on submitted scoring criteria, solutions, and point allocations. To serve as a control group, 14 additional instructor-authored questions and solutions were collected from publicly available instructional materials for which sources are presented in Appendix \ref{ax:online_sources_desc}. Their intended AI resistance is unknown. The research team also developed grading rubrics for these questions using their stated point allocations. All rubrics were normalized to a 100-point scale. The final dataset thus comprised 75 questions including 61 (81\%) faculty-contributed and 14 (19\%) from publicly available instructor-authored assessments.  
Table~\ref{tab:discipline_distribution} reports the distribution of questions across disciplines.

\end{minipage}
\hfill
\begin{minipage}[t]{0.38\columnwidth}
\vspace{0pt}
\centering
\scriptsize
\captionof{table}{\centering Distribution of the question dataset ($n=75$) across academic disciplines.}
\label{tab:discipline_distribution}
\setlength{\tabcolsep}{2pt}
\renewcommand{\arraystretch}{0.90}
\begin{tabular}{lccc}
\toprule
Discipline & \makecell{LLM-proof\\ $n$} & \makecell{Online\\ $n$} & \% \\
\midrule
Computer Science & 33 & & 44.0\% \\
Media Studies & & 10 & 13.3\% \\
Engineering & 5 & & 6.7\% \\
Statistics & 5 & & 6.7\% \\
Mathematics & 4 & & 5.3\% \\
Humanities & 3 & & 4.0\% \\
Psychology & 1 & 2 & 4.0\% \\
Physics & 2 & & 2.7\% \\
Politics & & 2 & 2.7\% \\
\makecell[l]{Other (Accounting,\\
Communications, Criminal\\
Justice, Education, History,\\
Journalism, Spanish, Theology)}
& 8 & & 10.7\% \\
\midrule
Total & 61 & 14 & 100\% \\
\bottomrule
\end{tabular}

\end{minipage}

\medskip

\noindent \textbf{Phase 2: }In the second phase of the IRB-approved competition, students were asked to select one or more Phase 1  questions (from any discipline) and answer them using publicly available LLMs and any prompting strategies. This phase served to assess the effectiveness of LLMs in answering our set of ``LLM-proof'' assessment questions. Students were required to submit a PDF containing screenshots documenting their complete interactions with the LLMs used to generate their responses. In total, 31 students from 10 academic majors submitted 451 answers across the question set. Students' demographics are reported in Appendix \ref{ax:student_demographics}. All submissions were manually graded (against faculty-provided solutions and grading rubrics) by three members of the research team independently. For each question, an aggregate score was computed by averaging the grades received across all student submissions associated with that question. In parallel, individual student rankings were determined using cumulative scores calculated as the sum of grades across all responses submitted by the student. This student scoring framework incentivized participants to maximize answer quality and to attempt a larger number of questions.

Both phases included monetary incentives. The five instructors whose questions received the lowest average scores each received \$1{,}000. Student awards were based on cumulative scores across all submitted answers. The two highest-ranked student participants each received \$1{,}000, the next two each received \$750, one third-place student received \$500, and five additional students received \$200 consolation prizes.

\begin{figure*}[t]
\centering
\includegraphics[width=0.95\textwidth]{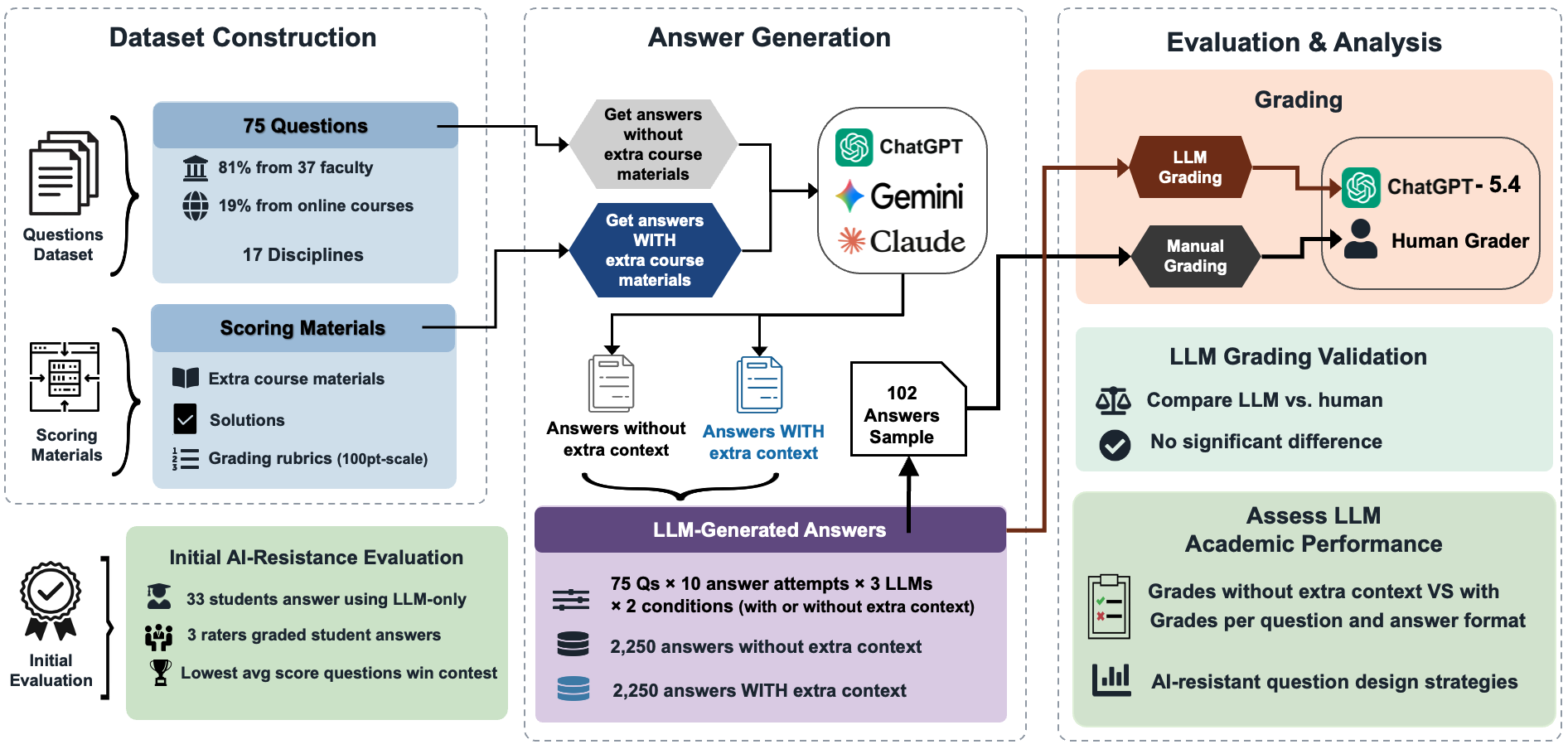} 
\caption{\small Overview of the methodology used to assess LLMs' performance on multidisciplinary instructor-crafted assessment questions}
\label{fig-cat_methodology}
\end{figure*}

\section{Generating LLM Answers}
\label{sec:llm_answers}
Following the construction of our 
``LLM-proof'' questions dataset, we completed a rigorous reproducible answer generation and evaluation process described in Figure \ref{fig-cat_methodology}. One limitation in our competition setup is that each student submitted only a single LLM-generated answer per question. Students were also not required to generate answers from different models. This setup failed to capture the variability in answers from a specific LLM, as well as the variability in answers across competing LLMs. These uncontrolled conditions produced an unbalanced corpus of generated answers. To establish consistent and reproducible findings, we generated multiple new answers to each instructor-authored question using different state-of-the-art LLMs. We also extended data collection by contacting participating faculty (from Phase 1) to obtain authentic course materials that students would typically use to answer the submitted questions. These materials included lecture notes, slides, and other instructional content intended to provide relevant contextual grounding for solving each assessment item. For the 14 questions sourced from publicly available instructional materials, we collected corresponding publicly available contextual materials. This additional context for each question enabled a controlled comparison of LLM performance with and without access to supplementary instructional materials. Using our dataset of questions combined with their solutions and grading rubrics, we describe below our procedure for generating a balanced corpus of LLM responses across models, both with and without supplementary course materials.

As illustrated in Figure \ref{fig-cat_methodology}, under the section \textit{LLM-Generated Answers}, for each of the 75 questions, we generated 10 responses without and 10 responses with extra course materials across three LLMs. This yields a total of 4{,}500 LLM-generated answers (75 questions $\times$ 10 answer attempts $\times$ 3 LLMs $\times$ 2 conditions) used throughout the analysis. The evaluated models consisted of the paid versions of ChatGPT-5.4, Gemini 2.5 Flash, and Claude Sonnet 4.6 (as seen in Figure \ref{fig-cat_methodology}, under the section \textit{Answer Generation}). 

\section{LLM-as-a-Judge for Grading}
\label{sec:llm_grading}
LLM-as-a-judge approaches have emerged as efficient alternatives to human grading, achieving comparable accuracy across disciplines, response formats, and grading strategies, with GPT-based models achieving up to 91.5\% on college-level tasks \cite{chen2025grading, kooli2025transforming, lee2024college, grevisse2024llm}.
We used GPT-5.4 as our grading model to score the 4{,}500 LLM-generated answers to questions. For each grading instance, the grading model was provided with the question, the LLM-generated answer, and the grading rubric associated with the question, an approach previously shown to improve grading reliability \cite{garcia2025chatgpt}. The prompt used for automated scoring of all answers is presented in Appendix \ref{ax:llm_prompt_scoring}. To assess the reliability of the automated grading pipeline, one of the authors manually graded a stratified sample of the dataset. This is illustrated in Figure \ref{fig-cat_methodology} under the section \textit{Grading}. Specifically, for one randomly selected question from each of the 17 academic disciplines, one response generated without additional course material and one response generated with additional course material were manually graded for each of the three LLMs. This procedure resulted in 102 manually graded responses (17 questions $\times$ 1 answer attempt $\times$ 3 LLMs $\times$ 2 conditions), corresponding to approximately 2\% of the total 4{,}500 LLM-generated answers. Human-graded scores to those 102 responses were subsequently compared against the grades produced by the GPT-5.4 grader for the same 102 responses. A paired $t$-test presented in Table~\ref{tab:human_vs_llm_grades} shows that differences between average LLM-graded scores and average human-graded scores remained relatively small across all evaluated conditions.

\noindent
\begin{minipage}[t]{0.57\columnwidth}
\vspace{0pt}
Without supplementary course material, the absolute mean-score differences were 3.7 points for ChatGPT (79.3 vs.\ 75.6), 7.5 points for Gemini (70.2 vs.\ 62.7), and 2.9 points for Claude (85.4 vs.\ 82.5) on a 100-point grading scale. With extra material, the differences narrowed to 2.1 points for GPT (83.7 vs.\ 81.6), 2.9 points for Gemini (74.5 vs.\ 71.6), and 6.2 points for Claude (82.8 vs.\ 76.6). Importantly, none of the paired comparisons yielded statistically significant differences ($p < 0.05$), suggesting that the automated grading process remained broadly aligned with human grading. These results support the use of GPT-5.4 as a reliable grading mechanism for our sample.
\end{minipage}
\hfill
\begin{minipage}[t]{0.40\columnwidth}
\vspace{0pt}
\centering
\small
\setlength{\tabcolsep}{5pt}
\renewcommand{\arraystretch}{0.7}
\captionof{table}{\centering \footnotesize Paired $t$-test comparing average human-graded and LLM-graded scores for 17 selected questions. Mean-L $=$ LLM mean score; Mean-H $=$ human mean score; \\ $\Delta =$ Mean-L $-$ Mean-H. No $p$-value is below 0.05.}
\label{tab:human_vs_llm_grades}
\begin{tabular}{@{}lrrrr@{}}
\toprule
LLM & Mean-L & Mean-H & $\Delta$ & $p$-value \\
\midrule
\multicolumn{5}{@{}l}{\textit{Without extra material ($n=17$)}} \\
GPT    & 79.3 & 75.6 & $+3.7$ & 0.537 \\
Gemini & 70.2 & 62.7 & $+7.5$ & 0.128 \\
Claude & 85.4 & 82.5 & $+2.9$ & 0.727 \\
\midrule
\multicolumn{5}{@{}l}{\textit{With extra material ($n=17$)}} \\
GPT    & 83.7 & 81.6 & $+2.1$ & 0.564 \\
Gemini & 74.5 & 71.6 & $+2.9$ & 0.407 \\
Claude & 82.8 & 76.6 & $+6.2$ & 0.196 \\
\bottomrule
\end{tabular}
\end{minipage}

\section{LLM Academic Performance}
\label{sec:rq1}
Leveraging LLM-as-a-judge framework, we analyze LLMs' performance to address \textbf{RQ1:How effective are state-of-the-art LLMs, when used to answer college-level assessment questions that are intentionally designed to be ``LLM-proof''?} We compute the mean scores achieved by each model across all 75 questions under both conditions (with and without supplementary course material). Figure~\ref{fig:llm_avg_grades} shows the results for each LLM. 

\noindent
\begin{minipage}[t]{0.52\columnwidth}
\vspace{0pt}
As seen in Figure~\ref{fig:llm_avg_grades}, without extra course material, all LLMs achieve consistently high average scores on both question sets (i.e., the 61 LLM-proof questions and 14 online-sourced questions). Each model scores higher on the online-sourced questions than on the LLM-proof questions (by 4.2 to 6.1 points, depending on the model). On the LLM-proof questions, GPT-5.4 achieves the highest mean score (83.2), followed by Claude Sonnet 4.6 (82.1) and Gemini 2.5 Flash (78.7). On the online-sourced questions, Claude Sonnet 4.6 ranks first (88.2), followed by GPT-5.4 (87.5) and Gemini 2.5 Flash (83.2). 

\end{minipage}
\hfill
\begin{minipage}[t]{0.43\columnwidth}
\vspace{0pt}
\centering
\includegraphics[width=\linewidth, height=4.0cm]{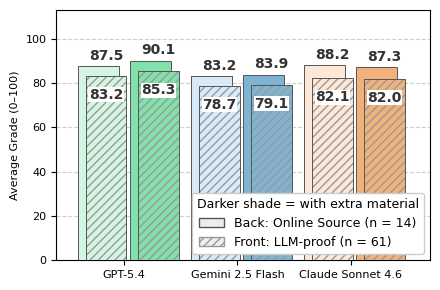}
\captionof{figure}{Distribution of LLM grades across three LLMs.}
\label{fig:llm_avg_grades}
\end{minipage}

\smallskip
Providing supplementary course material doesn't significantly impact LLMs' performance. GPT-5.4 only improves by $+2.1$ points (83.2 without vs.\ 85.3 with extra material) for LLM-proof questions and by $+2.6$ points for online-sourced questions (87.5 without vs.\ 90.1 with extra material). Gemini 2.5 Flash improves slightly on LLM-proof questions (by $+0.4$ points; 78.7 without vs.\ 79.1 with extra material) and online-sourced questions (by $+0.7$ points; 83.2 without vs.\ 83.9 with extra material). In contrast, Claude Sonnet 4.6 decreases slightly on LLM-proof questions (by $-0.1$ points; 82.1 without vs.\ 82.0 with extra material) and online-sourced questions (by $-0.9$ points; 88.2 without vs.\ 87.3 with extra material), indicating that additional material does not consistently translate into improved outcomes.

Overall, LLM performance on our moderately sized crowdsourced dataset of ``LLM-proof'' questions remains high across all three LLMs. This aligns with broader evidence that frontier LLMs can perform at or above passing thresholds on academic evaluations across disciplines without task-specific fine-tuning \cite{wang2024mmlu, achiam2023gpt, team2023gemini}. 
Within this scope, the results raise important concerns regarding the validity of assessment-based judgments of student learning, particularly in environments where GenAI is widely accessible. Higher education institutions face a growing need to reconsider how assessments are designed and how student learning is evaluated to ensure that academic performance remains a meaningful indicator of individual understanding and competence.

\section{Assessing Instructor-Tweaking Strategies}
Next, we examine question design strategies that are used by instructors to proactively increase assessment resistance to LLMs rather than relying on post-hoc detection of AI-generated work. 

\subsection{Faculty-Reported AI-Resistance Strategies}
\label{sec:tweaking_strategies}
When submitting their questions, faculty participants were asked to provide an explanation of what, in their view, made their questions resistant to AI systems. To address the first part of \textbf{\emph{RQ2-What types of strategies are used by instructors to create LLM proof assessments?}} we identify the themes in participants’ explanations by performing thematic analysis using an inductive approach \cite{clarke2017thematic}. Three of the authors independently reviewed the faculty written explanations in detail, capturing instructors’ perceptions of how AI resistance can be engineered through assessment design. After gaining initial insights, each of the three authors independently identified a set of themes. Next, all three authors collaboratively reviewed and refined their interpretations, compared independently identified themes, and consolidated overlapping themes through iterative deliberation. Discrepancies were resolved through consensus-building to establish the final themes underlying the strategies for improving AI resistance. Below, we highlight the five overarching strategies resulting from this thematic analysis.

\textbf{Strategy 1: Creating ``Unsearchable'' Original Content.}
Instructors using this strategy designed ``original'' problems from scratch (i.e., problems that did not draw upon any pre-existing material that is available on the Internet) so that solutions for these problems could not be found in any online sources and AI training datasets. Illustrating this proactive design choice, one instructor explained, ``this question was designed by me from scratch without any inspiration from the Internet'' (P3). Another shared that they ``just tried to create a question where students wouldn't be able to easily find solutions online'' (P22). Beyond preventing searchability, participants pointed out that tailored, non-standard problem structures would resist AI solution generation. P18 presented their submitted assessment item as ``an open-ended question that depends on the student's understanding of the subject; there's an infinite number of solutions.'' Similarly, P26 observed that ``the nature of [their] question makes it hard to answer correctly with AI''. Together, these accounts reflect instructors' attempts to reduce reliance on searchable answers by designing tasks that encourage demonstrations of students' own understanding and accommodate multiple possible solutions.

\textbf{Strategy 2: Exploiting Known AI Limitations.}
This theme captures strategies that explicitly target perceived weaknesses in LLM reasoning capabilities, especially in multimodal settings.
For example, a participant noted that their submitted math problem, which asked for the number of possible spellings of a given word, ``requires keeping track of the letters' positioning during the calculation steps which AI is notoriously bad at'' (P1). Another instructor expressed skepticism about ``LLMs’ abilities to do logical reasoning on top of image-based questions'' (P2). In addition to anticipating LLM weaknesses, some instructors pre-tested questions using LLMs and refined them based on observed performance. A couple of participants described ``running items through GPT'' to confirm low performance, noting mixed success across model versions (P4, P10), while another observed that certain parts of their questions were ``often incorrect when attempted by GenAI'' (P9). This iterative validation process reflects an adversarial design loop in which LLMs are used to probe and strengthen AI resistance.

\textbf{Strategy 3: Requiring Class-Specific Knowledge.}
This strategy involved embedding course- or institution-specific knowledge assumed to be unavailable in LLM training data or web resources. One participant explained that their question required students to discuss ``technical [Computer Science] concepts covered in class, using the notation used in class'' (P12), while another \emph{Computer Science} question included concepts that depended on class-specific assumptions where ``searching for it online or using GenAI may lead to using alternate assumptions'' (P34), which would lead to the incorrect answer. Other instructors emphasized that their questions relied on definitions unique to their course or textbook. For example, one mentioned that their question introduced a unique ``personal name for this method of solving [a math question] that AI can’t recognize'' (P23). One \emph{Communications} question required students to ``describe their decision-making process in alignment with the disaster check decision tree'', which was covered in class, but does not correspond to common terminology (P35). Several instructors also incorporated human-centered elements by requiring students to ``explain answers from individual perspectives'' (P6), or integrate ``personal memories'' into writing tasks (P33), reinforcing the view that human reasoning processes themselves contribute to AI resistance \cite{torkestani2025machine}. 

\textbf{Strategy 4: Adding Strategic Complexity.}
This broad theme encompasses multiple interrelated design strategies, including increasing cognitive load (e.g., by adding complexity through layered problem structures) and introducing deliberate misdirection in questions. Together, these approaches aim to make questions more resistant by requiring deeper structured understanding rather than surface-level pattern matching. 
For example, one instructor added ``adversarial noise to images'' to increase interpretive difficulty for an LLM (P7). Another explained  that ``while his original question had only integer addition, [he] changed it to get a more complex question that also includes integer multiplication'' (P14). 
Others mentioned explicitly ``embedding misdirection'' (P10) and adding confusion to get ``a significantly enhanced version of [originally] simpler problems'' (P28). Some instructors also  ``deleted guiding hints'' or removed explicit cues such as the ``problem name'' to increase difficulty for LLMs (P25).

\textbf{Strategy 5: Demanding Specific Output Formats.}
This last strategy involved requiring highly specific output formats that impose procedural and representational constraints beyond simple text generation. Instructors described requiring structured reasoning frameworks such as Issue--Rule--Application--Conclusion (IRAC) for a \emph{Journalism} question (P5), or requesting to ``draw k-maps and a block diagram'' for a \emph{Computer science} question (P27). An \emph{Industrial engineering} question required students to “draw a graph that describes the resulting distribution, and annotate how [they] interpret this question'' (P17). These constraints were viewed as increasing the need for domain-specific understanding and multimodal reasoning, thereby making questions more AI resistant.

\textbf{No Tweaking Strategy (Control Group).}
In addition to assessing instructors' ``LLM-proofing'' strategies, we incorporate a baseline control set to examine LLM performance on questions not specifically designed to resist AI systems. Out of the 75 questions included in the study, 61 were intentionally designed by course instructors to be resistant to LLM-generated answers using the strategies presented above, whereas the remaining 14 were collected from online courses, and their intended resistance to LLMs is unknown. These 14 questions serve as a control group providing a reference point for evaluating the effectiveness of instructors' assessment-design strategies in comparison to questions without an established LLM-resistance design objective.

\subsection{Comparing Tweaking Strategies}
\label{sec:tweaking_strategies_all_questions}
We now examine LLM performance across all 75 questions in our dataset, grouped by the tweaking strategy used to design them to address the second part of \textit{\textbf{RQ2: How effective are strategies used by instructors to create LLM proof assessments in constraining LLM performance?}}  
Each of the 61 faculty-submitted questions was assigned the strategy theme uncovered in Section \ref{sec:tweaking_strategies}. The remaining 14 questions had no reported tweaking strategy and are grouped under a \textit{None} category, which serves as a control group of questions not deliberately modified for AI resistance. 
For each of the 75 questions and each condition (with and without supplementary course materials), we compute the average score over 30 LLM-generated responses (10 answer attempts per LLM $\times$ 3 LLMs). We then compare the median of these per-question averages across strategies. 

As reported in Figure \ref{fig:tweaking_strategies}, LLMs achieved median scores above 75\% regardless of the tweaking strategy applied, and that raises questions about the sanctity of educational assessments in higher education. Further, Figure \ref{fig:tweaking_strategies} shows that as compared to the baseline questions (which achieve a median score of 88 without and 87 with course materials, respectively), questions modified through Strategy 1 (\textit{Create ``unsearchable'' original content}), Strategy 2 (\textit{Exploit known AI limitations}), and Strategy 3 (\textit{Require class-specific knowledge}) are slightly more AI-resistant as they achieve slightly lower median scores of 83, 84, and 82 without course materials, respectively.
Interestingly, LLM performance on these questions does not increase significantly with additional course material. Both Strategy 1 and Strategy 3 contained a few outlier questions with considerably lower scores (between 33 and 65), which suggests that if executed well, these strategies can be very effective at least for some specific questions. Strategy 4 (\textit{Add strategic complexity}) displayed the lowest median scores (78 without vs.\ 81 with course material), making this the strategy with a consistent reduction of roughly 7 to 10 points relative to the \textit{None} category. This group also includes the single most AI-resistant question in the dataset, a physics question on which LLM-generated answers averaged 3 out of 100 under both conditions.

\noindent
\begin{minipage}[t]{0.48\columnwidth}
\vspace{0pt}
In contrast, Strategy 5 (\textit{Demand specific output formats}) produced the highest median scores of all groups (91 without vs.\ 92 with course material), exceeding even the \textit{None} category with very little variability across questions. To explore this pattern further, Section \ref{sec:answer_types} examines LLM performance by answer type, assessing whether these results extend across the range of answer formats represented in our dataset.

Based on our results in Figure \ref{fig:tweaking_strategies}, Strategy 4 (\textit{Add strategic complexity}) appears to be the most AI-resistant strategy, as it consistently helps produce LLM-generated answers with lower scores. Of course, care must be taken not to over-interpret these results because of our small sample sizes. We note that this comparison is between different questions. Thus, differences between tweaking strategies also reflect variation in discipline and question difficulty. We address this limitation with a paired analysis of original and tweaked question variants in the next subsection.

\end{minipage}
\hfill
\begin{minipage}[t]{0.48\columnwidth}
\vspace{0pt}
\centering
\includegraphics[width=\linewidth, height=6.5cm]{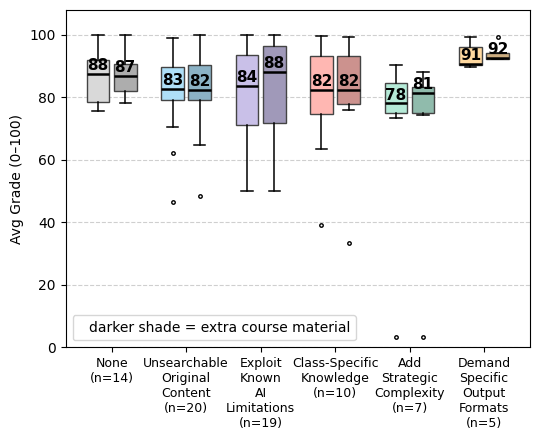}
\captionof{figure}{\centering  Comparing tweaking strategies used by instructors to make their assessment questions ``LLM-proof.'' n is the number of questions.}
\label{fig:tweaking_strategies}
\end{minipage}



\subsection{Analysis of Original and Tweaked Questions}
\label{sec:original_vs_tweaked}
Among the 61 questions submitted directly by faculty participants, 11 (18\%) questions represented tweaked versions of old (possibly non-AI resistant) questions. For such questions, faculty participants were asked to include both the original question and the ``tweaked'' variant specifically designed to increase resistance to LLM-generated answers. These paired submissions provide a unique opportunity to examine whether the modifications themselves, rather than differences in discipline or question difficulty, drive the patterns described in Section \ref{sec:tweaking_strategies_all_questions}.
To enable the comparison between each tweaked question with its own original, we generated responses for the 11 original question variants and scored them using the same pipeline applied to the main dataset (4{,}500 LLM-generated answers in Figure \ref{fig-cat_methodology}). This included identical generation settings,  multiple answer attempts, LLM-as-a-Judge grading procedures, and rubric-based scoring criteria described in Sections \ref{sec:llm_answers} and \ref{sec:llm_grading}.

The 11 paired questions cover four of the five strategies described in Section \ref{sec:tweaking_strategies}, specifically, Strategy 2 (\textit{Exploit known AI limitations}, $n=1$), Strategy 3 (\textit{Require class-specific knowledge}, $n=1$), Strategy 4 (\textit{Add strategic complexity}, $n=5$), and Strategy 5 (\textit{Demand specific output formats}, $n=4$). For each question variant and condition (with or without supplementary course materials), we compute the average score across 10 answer attempts for each LLM and calculate the median of the three resulting per-LLM averages.
Table \ref{tab:original_vs_tweaked_scores} reports these medians for the original and tweaked variants and their difference ($\Delta = \text{tweaked} - \text{original}$). A negative $\Delta$ indicates a lower median score for the tweaked variant, suggesting increased resistance to LLM-generated answers. 

Consistent with the findings reported in Section \ref{sec:tweaking_strategies_all_questions}, Table \ref{tab:original_vs_tweaked_scores} shows that Strategy 4 (\textit{Add strategic complexity}) yielded the most consistent reductions in LLM performance. Four of the five questions modified using this strategy had lower median scores in the tweaked variant under both conditions, with decreases ranging from 0.5 to 9.2 points.

\begin{minipage}[t]{0.42\columnwidth}
\vspace{0pt}
\setlength{\parindent}{0pt}
The largest decrease was observed for a computer science question (\emph{CompSci\_30}), for which the LLM median score dropped from 95 to 87 without course materials and from 95 to 86 with them. This pattern provides additional, albeit limited, evidence that adding strategic complexity may be an effective approach to increasing the AI resistance of college-level assessment questions. In contrast, Strategy 5 (\textit{Demand specific output formats}) produced mostly small differences or higher scores for tweaked variants. For one of the questions tweaked using this strategy, the median score increased from 84 to 96 without course materials and from 88 to 93 with them; the other three showed differences of less than 3 points in either direction. Consistent with Section \ref{sec:tweaking_strategies_all_questions}, these findings suggest that constraining output format alone may not substantially impair LLM performance.

\end{minipage}
\hfill
\begin{minipage}[t]{0.52\columnwidth}
\vspace{0pt}
\centering
\fontsize{7pt}{9.5pt}\selectfont
\setlength{\tabcolsep}{4pt} 
\renewcommand{\arraystretch}{0.92}
\captionof{table}{\centering Original vs tweaked question median scores. \\ TM $=$ Tweaked Median. OM $=$ Original Median. $\Delta M = \text{TM} - \text{OM}$. \\ \textbf{Values in Bold Text} $=$ negative $\Delta M$ (harder for LLMs).}
\label{tab:original_vs_tweaked_scores}
\begin{tabular}{@{} l r r @{} r c r r @{} r @{}}
\hline
& \multicolumn{3}{c}{\textbf{Without Extra Material}} & & \multicolumn{3}{c}{\textbf{With Extra Material}} \\
\cline{2-4} \cline{6-8}
\textbf{Question} & \textbf{OM} & \textbf{TM} & \textbf{$\Delta M$} & & \textbf{OM} & \textbf{TM} & \textbf{$\Delta M$} \\
\hline
\multicolumn{8}{@{}l}{Strategy 2: \textit{Exploit known AI limitations}} \\
Mathematics\_01 & 60.0 & 52.6 & \textbf{-7.4} & & 40.0 & 68.8 & 28.8 \\
\hline
\multicolumn{8}{@{}l}{Strategy 3: \textit{Require class-specific knowledge}} \\
CriminalJustice\_01 & 53.6 & 87.6 & 34.0 & & 66.8 & 84.6 & 17.8 \\
\hline
\multicolumn{8}{@{}l}{Strategy 4: \textit{Add strategic complexity}} \\
Accounting\_01 & 77.1 & 81.0 & 3.9 & & 77.8 & 79.8 & 2.0 \\
CompSci\_08 & 79.4 & 75.5 & \textbf{-3.9} & & 78.8 & 75.6 & \textbf{-3.2} \\
CompSci\_30 & 94.9 & 87.4 & \textbf{-7.5} & & 95.2 & 86.0 & \textbf{-9.2} \\
CompSci\_33 & 75.8 & 74.4 & \textbf{-1.4} & & 76.8 & 76.3 & \textbf{-0.5} \\
Theology\_01 & 80.4 & 78.0 & \textbf{-2.4} & & 83.2 & 81.6 & \textbf{-1.6} \\
\hline
\multicolumn{8}{@{}l}{Strategy 5: \textit{Demand specific output formats}} \\
CompSci\_02 & 91.4 & 88.7 & \textbf{-2.7} & & 90.1 & 92.0 & 1.9 \\
CompSci\_29 & 83.7 & 96.2 & 12.5 & & 88.1 & 93.0 & 4.9 \\
Engineering\_01 & 98.6 & 99.4 & 0.8 & & 98.9 & 99.0 & 0.1 \\
Engineering\_04 & 87.0 & 89.4 & 2.4 & & 91.0 & 90.4 & \textbf{-0.6} \\
\hline
\end{tabular}
\end{minipage}

\smallskip

Although the paired analysis is based on a small sample, its results broadly align with the findings from the comparison of all five tweaking strategies across all questions. They also indicate that modifications intended to increase difficulty may not always have the intended effect and may, in some cases, inadvertently make assessment questions more accessible to LLMs. Instructors should therefore test their tweaked questions on state-of-the-art LLMs before releasing them to their students as course assessment items. 



\section{Examining Answer Types}
\label{sec:answer_types}
The 61 ``LLM-proof'' assessment questions examined in this study differ not only in their content but also in the form of the answer they require, ranging from written explanations to step-by-step calculations. Based on these differences, we now try to answer \textit{\textbf{RQ3: What characteristics of assessment questions, if any, affect the ability of LLMs to answer them?}} To investigate this question, we classify each of the 61 ``LLM-proof'' assessment questions into one of four answer-format categories identified in the dataset, capturing differences in the expected form of the response. 

The first category  \textbf{Reasoning steps / calculation} ($n{=}30$) contains questions that require step-by-step derivations, formal proofs, or structured mathematical computations. The second, \textbf{Programming} ($n{=}13$), encompasses questions that require writing computer code or describing an appropriate software architecture. The third category, \textbf{Written expression / case study analysis} ($n{=}12$) covers questions which expect open-ended text as an answer, including essays, reflective writing such as prayers, content generation for web pages, or interpreting a provided scenario. Lastly, \textbf{Graphical representation} ($n{=}6$) questions require the construction of conceptual diagrams or figures as an answer.  This categorization builds on prior work evaluating LLM performance across categories such as reasoning, symbolic manipulation, and open-ended generation tasks \cite{srivastava2023beyond, wang2024measuring}. 

All scores presented in this analysis are from the main dataset of LLM-generated answers, using the generation and grading procedures described in Sections \ref{sec:llm_answers} and \ref{sec:llm_grading}. For each of the 61 questions and each condition (with and without supplementary course materials), we compute the average score over 30 LLM-generated responses (10 answer attempts per LLM $\times$ 3 LLMs). We then compare the median of these per-question averages across the four answer types. 
As reported in Figure \ref{fig:answer_formats}, LLMs achieved high scores for all answer formats, with small differences between formats. Questions requiring \textit{Reasoning steps / calculation}, the most common format in our dataset, reached a median score of 83 without and 82 with course material.  

\noindent
\begin{minipage}[t]{0.57\columnwidth}
\vspace{0pt}
As seen in Figure \ref{fig:answer_formats}, this group also showed considerable variability across questions, with interquartile ranges of about 22 points since it contains the most AI-resistant question in the dataset, a physics question on which LLM-generated answers averaged 3 out of 100 under both conditions. This performance variability on tasks requiring step-by-step reasoning and mathematical calculations is an expected limitation of LLMs, including reasoning models such as those examined in our study \cite{openai2026gpt5-4, google2025start, anthropic2026sonnet4-6}. Prior work has demonstrated limited model robustness on tasks requiring multi-step logical reasoning, showing that seemingly irrelevant perturbations, such as variations in number parsing, can disrupt reasoning \cite{nezhurina2024alice}.

\end{minipage}
\hfill
\begin{minipage}[t]{0.40\columnwidth}
\vspace{0pt}
\centering
\includegraphics[width=\linewidth, height=4.5cm]{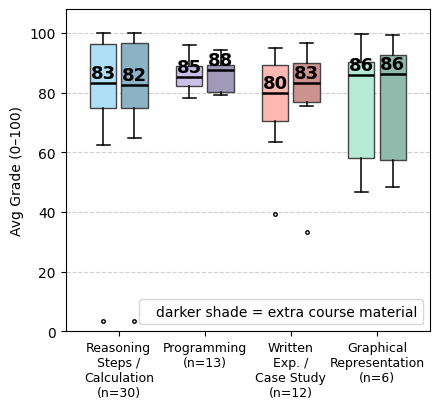}
\captionof{figure}{\footnotesize \centering LLM performance on different answer types.}
\label{fig:answer_formats}
\end{minipage}

\smallskip
\textit{Programming} questions reached a median of 85 without and 88 with course material, and showed the least variability of all formats. This suggests that code generation and software-oriented reasoning are comparatively well-supported capabilities in current LLMs, consistent with prior evaluations showing strong performance in code synthesis and debugging tasks under constrained specifications \cite{nam2024using, chen2021evaluating}. Of course, we anticipate these numbers to be even higher with Codex and Claude Code models \cite{pinna2026comparing}, but those were not the focus of our analysis. Lastly, \textit{Graphical representation} questions had a median of 86 under both conditions, but showed the largest variability of all formats, with interquartile ranges of 32 to 35 points. Two of the six questions in this group, both from the humanities, averaged only 47 to 50. Although this group is small, this pattern suggests that graphical answer formats do not systematically limit LLM performance, but may do so for specific questions. Our findings align with existing studies observing inconsistent performance from visual LLMs in comparison to text-only LLMs when answering college-level academic questions \cite{yue2024mmmu}

\section{Discussion \& Conclusion}
This study examined the extent to which state-of-the-art LLMs can correctly answer instructor-crafted ``LLM-proof'' higher-education assessment questions spanning 17 academic disciplines, 11 institutions, and diverse response formats. Through thematic analysis, we identified 5 distinct strategies used by instructors to make their assessment items resistant to AI cheating. Our findings show that while one strategy (Adding Strategic Complexity in Questions) is effective at hindering LLM performance on questions, the other strategies do not create a significant dent in LLM performance. Importantly, across the three research questions, the results suggest that LLMs are capable of producing high-scoring responses on authentic college-level assessment questions, even when the questions are intentionally designed to resist AI-assisted completion. The strongest-performing model, GPT-5.4 achieved an 83.2\% average score without supplementary instructional material on our set of LLM-proof questions and 87.5\% on online-sourced questions. Together, all of our findings seem to indicate that instructor-crafted college-level assessments across disciplines are increasingly vulnerable to unauthorized AI assistance, and while strategies used by instructors to LLM-proof their assessments are somewhat effective, these strategies do not represent a silver bullet by themselves. In parallel, instructors and higher-ed institutions should be proactively innovating the delivery of courses on college campuses, so as to remain relevant in an era where GenAI use (and abuse) is ubiquitous. In the following paragraphs, we discuss important implications for the design of higher-education assessment, academic-integrity policy, and the broader ethics of deploying LLMs in educational settings.

\noindent \textbf{Implications for Assessment Design.} 
Our findings reveal a widening gap between the increasing capabilities of LLMs and instructors' ability to design AI-resistant assessments. This challenge is compounded by the limitations of conventional plagiarism detection systems and commercial AI detectors in identifying AI-generated content \cite{lim2023generative}. Although the scope of our study is limited, the results raise important concerns about the continued feasibility of relying on take-home assignments and online examinations as reliable indicators of students' independent understanding, particularly given the growing availability of LLMs \cite{rudolph2023chatgpt, walsh2025everyone}. Our findings in Sections \ref{sec:tweaking_strategies_all_questions} \& \ref{sec:original_vs_tweaked} suggest that adding strategic complexity to assessment questions offers a potential avenue for increasing the AI-resistance of academic tasks. At the same time, our findings suggest that tweaking the expected output formats of assessment questions does not lead to effective ``LLM-proofing'' of assessment questions. 

However, rather than relying exclusively on increasingly complex tasks or questions intended to be AI-resistant, instructors may need to reconsider how learning is assessed and demonstrated. For example, a more constructive adaptation of course assessments may lie in the deliberate integration of LLMs into assessment practice while preserving meaningful demonstrations of student understanding. Students might, for instance, be required to complete tasks in tandem with an LLM working as a teammate \cite{lyu2025will}, or to critically analyze and refine model-generated responses to a given prompt. Such assessment designs shift the focus from merely producing answers (which LLMs can do with ease, as evidenced by our results in Section \ref{sec:rq1}) to demonstrating genuine intellectual effort and informed human oversight of AI systems. At the same time, institutions need to invest in initiatives that help instructors redesign their courses and assessments to reflect the ever-evolving landscape of GenAI tools and establish discipline-specific AI guidelines.

\noindent \textbf{Implications for Institutional Policies.}  Existing integrity frameworks were largely developed for forms of misconduct involving plagiarism, contract cheating, or unauthorized collaboration \cite{dwivedi2023opinion}. As students continuously find new ways to submit undetected AI-generated work \cite{thunder2025white}, the long-term effectiveness of current approaches in the enforcement of academic-integrity principles is uncertain, particularly when institutional policies regarding permissible AI use are absent or provide ambiguous guidance \cite{thacker2026states, smit2025ambiguous}.
In this new context, institutions may need to transition from predominantly prohibition-based approaches toward governance practices that support the integration of LLMs into the educational pipeline. For example, as opposed to negatively judging AI use as ``cheating'' or an ``academic integrity violation'' which needs to be tackled via punitive action, institutions may consider engaging instructors across disciplines to help promote an environment of acceptance around AI use, as long as it is accompanied by appropriate disclosure of what AI was used for (and what it was not used for). Instructors need to lead the way, e.g., by disclosing to their students any time they use GenAI to make exam questions, slides, figures, or other instructional material in their courses, or if they use GenAI to grade student homework/exam copies. At the same time, new courses can be designed in institutions where students are explicitly taught about the importance (and process) of honest self-disclosure around AI use. Over time, this transition would reduce the need for institutions to actively police unauthorized AI usage and focus on promoting transparency and disclosure of AI assistance \cite{dwivedi2023opinion, corbin2025talk}. The recent story about the Provost of Dartmouth College is a welcome example in this direction, where the Provost was asked to resign not because they used AI, but because they failed to disclose AI use in their academic work \cite{apnews2026darmouth}.

In parallel, institutions should also reconsider what academic achievement and degree completion signify in an era of increasingly capable LLMs. Our findings in Section \ref{sec:rq1} raise concerns that students may be able to obtain passing grades on course assessments through LLM assistance without needing to attend classes, and without developing the underlying knowledge and competencies those assessments are intended to measure. This possibility challenges the core business model of higher-education institutions, i.e., students ``need'' to attend in-person course instruction via human instructors regularly to successfully achieve good grades. Therefore, for the long-term viability of institutions, it is imperative that they should explore incentive structures that promote consistent classroom attendance, active participation, and independent student engagement with coursework, ensuring that students remain meaningfully involved in their own learning.

\noindent \textbf{Implications for LLM Developers.} This study additionally contributes to ongoing ethical debates surrounding the deployment of LLMs in educational environments, particularly with the introduction of tools such as ChatGPT Edu \cite{openai2024edu}. While LLMs may enhance productivity and personalized support, they also raise concerns regarding cognitive offloading, overreliance, and the erosion of foundational learning skills \cite{lee2025impact, gerlich2025ai}. Ultimately, addressing these challenges requires shared responsibility between educational institutions and the companies developing LLMs. Developers should consider incorporating educationally informed guardrails into LLMs that prioritize encouraging critical thinking rather than readily supplying answers to coursework assessments. Ultimately, a constructive response to these challenges requires the development of assessment ecosystems and AI tools that meaningfully integrate generative AI while preserving the independent intellectual development that higher education is intended to cultivate.

\noindent \textbf{Equity Considerations.} More broadly, our findings raise important equity considerations. Access to high-performing LLMs is uneven across socioeconomic, geographic, linguistic, and institutional contexts \cite{zhang2026college, bassignana2025ai}. Students with access to subscription-based models, advanced prompting expertise, or greater technological resources may gain disproportionate advantages \cite{bencsik2026gets} in educational assessment environments where AI usage is difficult to monitor. These disparities risk exacerbating existing educational inequalities \cite{holmes2022ethics}. Furthermore, students who avoid using LLMs for ethical or financial reasons \cite{beckman2025genai} may become structurally disadvantaged relative to peers who extensively integrate these systems into coursework. Consequently, education regulatory authorities face not only an academic integrity challenge but also a fairness challenge in determining how AI usage should be regulated, disclosed, or incorporated into learning outcomes.

\section{Limitations and Future Work}
This study has limitations that shape the interpretation and generalizability of its findings. First, although the dataset spans 17 academic disciplines and 11 institutions, the number of assessment questions is relatively modest. The dataset was constructed from authentic instructor-crafted college assessments, accompanied by solutions, grading rubrics, and faculty explanations, which required substantial instructor participation and manual curation. While this approach captures variation in real-world assessment practices across disciplines and answer formats, the limited number of questions constrains the extent to which the results can be generalized to college assessments more broadly.


Second, while the dataset includes heterogeneous question formats and disciplines, the distribution of questions across categories remains uneven. Certain disciplines are represented more heavily than others, and some answer-format categories contain relatively few observations. This imbalance limits the statistical power of comparisons across certain subgroups and suggests caution when generalizing fine-grained conclusions regarding specific assessment types. 

Future research should therefore expand both the scale and scope of multidisciplinary instructor-crafted assessment evaluation to explore whether the observed patterns hold across a wider range of educational contexts. One important direction involves investigating how students strategically integrate LLMs into coursework, including the relationship between prompting expertise, socioeconomic access to LLMs, and educational outcomes.

\section*{Generative AI Use Disclosure Statement} 
Generative AI tools were used to support manuscript preparation and data visualization. ChatGPT-6 was used for grammatical refinement and to restructure content for conciseness, ensuring compliance with page limit requirements, but each LLM-generated rephrase was manually verified by the authors to ensure that it represents exactly the message that the authors wanted to convey. Claude Opus-5.5 assisted with generating Python scripts in Jupyter Notebook environments to produce the figures and tables included in the manuscript. However, the authors manually read and verified the correctness of the generated Python scripts before running them to generate the figures/tables in the paper.

\section*{Ethical Statement}
This study raises important ethical considerations associated with the use of GenAI in education-related research. One central concern involves the sharing of instructor-crafted assessment materials with commercial large language models. Because many frontier LLM providers retain user interactions for system improvement or future model training, submitting authentic assessment questions to these systems may create potential privacy and intellectual-property risks. In particular, there exists a possibility that proprietary or previously unpublished assessment materials could become indirectly incorporated into future model training datasets. We acknowledge this concern as an important ethical consideration of conducting empirical evaluations using commercial generative AI systems.

At the same time, evaluating how frontier LLMs perform on instructor-authored assessments necessarily requires interacting with those systems using authentic educational materials. Understanding the extent to which current college-level assessments remain vulnerable to unauthorized AI-assisted completion is critical for informing educational policy, assessment redesign approaches, and responsible institutional responses to generative AI. All participating faculty members voluntarily contributed their assessment materials and provided informed consent for their use in this research study. The study was conducted under institutional review board (IRB) approval, and faculty participants were informed about the potential risks associated with sharing assessment content with third-party AI systems.

\begin{acks}
We sincerely thank all faculty participants who generously contributed self-authored assessment questions, solutions, grading rubrics, and supplementary instructional materials to this study. Their willingness to share authentic assessment materials and reflect critically on the challenges posed by generative AI made this research possible. 
\end{acks}

\bibliographystyle{ACM-Reference-Format}
\bibliography{references}

\newpage
\appendix
\label{sec:appendix}

\section{Faculty Participants Submission Form}
\label{ax:faculty_submission_form}
The following questions comprised the faculty submission form used to collect faculty-crafted assessment questions for this study.

\begin{enumerate}[leftmargin=0.9cm, topsep=0pt, itemsep=0pt, parsep=0pt]

  \item[\textbf{Q1}] Enter your faculty \texttt{.edu} email address.

  \item[\textbf{Q2}] What university or college are you affiliated with?

  \item[\textbf{Q3}] What department are you affiliated with?

  \item[\textbf{Q4}] What best describes your role?

  \item[\textbf{Q5}] In what discipline is the question (from an exam or assignment) that you will submit? Select one of the options provided.

  \item[\textbf{Q6}] If ''Other'' discipline selected, please specify. Otherwise enter ''NA''.

  \item[\textbf{Q7}] Upload a single question (from an exam, assignment, or project) that you believe is hard to solve with GenAI. 
  \textit{(Upload a single PDF file.)}

  \item[\textbf{Q8}] Upload the original version of your question, if you made any tweaks for the question to be hard to solve with GenAI. 
  \textit{(Upload a single PDF file.)}

  \item[\textbf{Q9}] Upload the solution (i.e., the ideal answer) that gets full points for your question, and add a rubric in the question below. 
  \textit{(Upload a single PDF file.)}

  \item[\textbf{Q10}] If available, upload a scoring rubric to facilitate the grading of your question. 
  \textit{(Upload a single PDF file.)}

  \item[\textbf{Q11}] Describe how you tweaked or designed the question (from your exam, assignment, or project) to make it hard to solve with GenAI.

  \item[\textbf{Q12}] Complete this 5-minute survey to share your experience.

\end{enumerate}

\newpage
\section{Faculty Participants Demographics}
\label{ax:faculty_demographics}
We present the demographics of our faculty participants in the table below.

\begin{table}[H]
\centering
\small
\begin{tabular}{ll}
  \toprule
  \textbf{Characteristic} & \makecell{\textbf{Count (n)}} \\
  \midrule
  \textbf{Role} & \\
  \quad Assistant Professor  & 11 \\
  \quad Associate Professor  & 9  \\
  \quad Other Faculty        & 5  \\
  \quad Professor            & 4  \\
  \quad Teaching Professor   & 3  \\
  \quad Adjunct Professor    & 1  \\
  \quad Not reported         & 4  \\
  \midrule
  \textbf{Department}        & \\
  \quad Computer Science              & 9 \\
  \quad Information Sciences          & 5 \\
  \quad Engineering                   & 5 \\
  \quad Humanities \& Social Sciences & 5 \\
  \quad Mathematics \& Statistics     & 2 \\
  \quad Business Management                     & 2 \\
  \quad Education                     & 2 \\
  \quad Physics                       & 1 \\
  \quad Library Science               & 1 \\
  \quad Not reported                  & 5 \\
  \midrule
  \textbf{Age}               & \\
  \quad 18 to 24             & 5  \\
  \quad 25 to 34             & 7  \\
  \quad 35 to 44             & 15 \\
  \quad 45 to 54             & 7  \\
  \quad 55 and over          & 3  \\
  \midrule
  \textbf{Gender}            & \\
  \quad Male                 & 24 \\
  \quad Female               & 10 \\
  \quad Prefer not to say    & 3  \\
  \midrule
  \textbf{Generative AI Usage} & \\
  \quad Never                                      & 3  \\
  \quad Rarely (less than once a month)            & 2  \\
  \quad Occasionally (1--3 times a month)          & 12 \\
  \quad Frequently (once a week)                   & 4  \\
  \quad Very frequently (multiple times per week)  & 9  \\
  \quad Daily                                      & 7  \\
  \bottomrule
\end{tabular}
\caption{Demographics of faculty participants ($N = 37$).}
\label{tab:demographics}
\end{table}

\bigskip

\section{Student Participants Demographics}
\label{ax:student_demographics}
We present the demographics of our student participants in the table below.

\begin{table}[H]
\centering
\small
\begin{tabular}{ll}
  \toprule
  \textbf{Characteristic} & \makecell{\textbf{Count (n)}} \\
  \midrule
  \textbf{Role} & \\
  \quad Undergraduate student  & 17 \\
  \quad Graduate student       & 13 \\
  \quad Other                  & 1  \\
  \midrule
  \textbf{Department}        & \\
  \quad Engineering                        & 7 \\
  \quad Computer Science                   & 5 \\
  \quad Information Sciences \& Technology & 5 \\
  \quad Business                           & 3 \\
  \quad Science \& Statistics              & 3 \\
  \quad Earth \& Mineral Sciences          & 2 \\
  \quad Health \& Human Development        & 2 \\
  \quad Division of Undergraduate Studies  & 2 \\
  \quad Education                          & 1 \\
  \quad Communications                     & 1 \\
  \midrule
  \textbf{Age}               & \\
  \quad 18 to 24             & 19 \\
  \quad 25 to 34             & 9  \\
  \quad 35 to 44             & 3  \\
  \midrule
  \textbf{Gender}            & \\
  \quad Male                 & 23 \\
  \quad Female               & 7  \\
  \quad Prefer not to say    & 1  \\
  \midrule
  \textbf{Generative AI Usage} & \\
  \quad Rarely (less than once a month)            & 4  \\
  \quad Occasionally (1--3 times a month)          & 1  \\
  \quad Frequently (once a week)                   & 4  \\
  \quad Very frequently (multiple times per week)  & 13 \\
  \quad Daily                                      & 9  \\
  \bottomrule
\end{tabular}
\caption{Demographics of student participants ($N = 31$).}
\label{tab:student-demographics}
\end{table}

\newpage
\section{Online-Sourced Materials Description}
\label{ax:online_sources_desc}
\begin{table}[H]
\centering
\small
\begin{tabular}{l>{\centering\arraybackslash}p{5cm}>{\centering\arraybackslash}p{7cm}}
  \toprule
  \textbf{Question} & \textbf{Online Source} & \textbf{Supplementary Course Material} \\
  \midrule
  MediaStudies\_01 & \href{https://www.wjec.co.uk/media/bzlhh4lk/wjec-gcse-media-studies-sams-from-2017-e.pdf}{WJEC UK Sample Assessment Materials}\tablefootnote{\url{https://www.wjec.co.uk/media/bzlhh4lk/wjec-gcse-media-studies-sams-from-2017-e.pdf}}  & \textit{The Media Student's Book} (5th ed.),  \cite[Chapter 3.9]{branston2010media}\\
  MediaStudies\_02 & \href{https://www.wjec.co.uk/media/bzlhh4lk/wjec-gcse-media-studies-sams-from-2017-e.pdf}{WJEC UK Sample Assessment Materials} & \cite[Chapter 2.12]{branston2010media}\\
  MediaStudies\_03 & \href{https://www.wjec.co.uk/media/bzlhh4lk/wjec-gcse-media-studies-sams-from-2017-e.pdf}{WJEC UK Sample Assessment Materials} & \cite[Chapter 1.1]{branston2010media} \\
  MediaStudies\_04 & \href{https://www.wjec.co.uk/media/bzlhh4lk/wjec-gcse-media-studies-sams-from-2017-e.pdf}{WJEC UK Sample Assessment Materials} & \cite[Chapter 11.1]{branston2010media}\\
  MediaStudies\_05 to 10 & \href{https://www.wjec.co.uk/media/bzlhh4lk/wjec-gcse-media-studies-sams-from-2017-e.pdf}{WJEC UK Sample Assessment Materials} & \cite[Chapter 14]{branston2010media}\\
  Politics\_01     & \href{https://apcentral.collegeboard.org/media/pdf/ap24-apc-us-go-po-q2-set-2.pdf}{AP Central College Board Sample 1}\tablefootnote{\url{https://apcentral.collegeboard.org/media/pdf/ap24-apc-us-go-po-q2-set-2.pdf}} & \textit{We the people: an introduction to American politics} (11th ed.),  \cite[Chapter 10]{ginsberg2021we}     \\
  Politics\_02     & \href{https://apcentral.collegeboard.org/media/pdf/ap24-apc-us-go-po-q1-set-2.pdf}{AP Central College Board Sample 2}\tablefootnote{\url{https://apcentral.collegeboard.org/media/pdf/ap24-apc-us-go-po-q1-set-2.pdf}} & \cite[Chapter 12]{ginsberg2021we} \\
  Psychology\_02   & \href{https://moodle.cloud.triadhq.com/}{Triad HQ Sample 1}\tablefootnote{\url{https://moodle.cloud.triadhq.com}} & \textit{Ethics in psychology: Professional standards and cases} (Vol. 3),  \cite[Chapter 17, P. 462-5]{koocher1998ethics} \\
  Psychology\_03   & \href{https://moodle.cloud.triadhq.com/}{Triad HQ Sample 2} & \textit{Cognitive psychology: A student's handbook},  \cite[Chapter 10, P. 498-501]{eysenck2013cognitive}\\
  \bottomrule
\end{tabular}
\caption{Online-sourced questions ($n = 14$), with their online sources and extra course material.}
\label{tab:online-sourced-questions}
\end{table}

\newpage
\section{Prompts Used for Answer Generation}
\label{ax:llm_prompts}

During the answer generation process, the temperature parameter was fixed at 1.0 for all LLMs and all generations to promote response diversity while maintaining a consistent generation configuration across models. Prior work and provider documentation indicate that higher temperature settings increase output variability while preserving overall response coherence \cite{openai2026create}, making this configuration appropriate for sampling-based evaluations intended to capture diverse model behaviors. The prompts used for answer generation with and without extra course material are presented in Appendices \ref{ax:prompt_answer-without-context} and \ref{ax:prompt_answer-with-context}.

The prompts below are reproduced verbatim. Answer-generation prompts presented are used for all three LLMs evaluated in our study, namely GPT-5.4, Gemini 2.5 Flash, and Claude Sonnet 4.6.

\subsection{Answer Generation Without Supplementary Materials}
\label{ax:prompt_answer-without-context}

This prompt is used to generate a student-style answer to a faculty-authored question provided as a PDF, without supplementary reference material.

\begin{quote}
\ttfamily\small\raggedright
\setlength{\parskip}{0.5\baselineskip}

You are an expert student aiming for the highest possible grade.

The academic question is provided as an attached PDF. Read it carefully,
including any diagrams, figures, or graphics, then write a thorough,
well-structured answer.

Requirements:
\begin{itemize}[noitemsep, topsep=2pt]
  \item[-] Address every part of the question completely.
  \item[-] Show your reasoning step by step where appropriate.
  \item[-] Use clear academic prose. Equations and bullet points are
    welcome where they aid clarity.
  \item[-] Do not say you are an AI. Just answer the question as a
    top student would.
\end{itemize}

\end{quote}

\subsection{Answer Generation With Supplementary Materials}
\label{ax:prompt_answer-with-context}

This prompt is used when one or more supplementary materials in PDF format are provided in addition to the instructor-authored question.

\begin{quote}
\ttfamily\small\raggedright
\setlength{\parskip}{0.5\baselineskip}

You are an expert student aiming for the highest possible grade.

The academic question is provided as an attached PDF, along with one or
more PDFs of extra reference material. Read all of them carefully,
including any diagrams, figures, or graphics.

Requirements:
\begin{itemize}[noitemsep, topsep=2pt]
  \item[-] Address every part of the question completely.
  \item[-] Refer to the extra material wherever it is relevant. Cite
    specific facts, figures, or arguments from it to strengthen your answer.
  \item[-] Show your reasoning step by step where appropriate.
  \item[-] Use clear academic prose. Equations and bullet points are
    welcome where they aid clarity.
  \item[-] Do not say you are an AI. Just answer the question as a
    top student would.
\end{itemize}

\end{quote}

\section{Prompt Used for Automated Scoring}
\label{ax:llm_prompt_scoring}

The prompt below is reproduced verbatim. It was used to generate scores using GPT-5.4. The same prompt was used to score every answer generated by all three LLMs evaluated in our study, namely GPT-5.4, Gemini 2.5 Flash, and Claude Sonnet 4.6.

\begin{quote}
\ttfamily\small\raggedright
\setlength{\parskip}{0.5\baselineskip}

You are a strict but fair examiner.

You are given: 1.~A student's answer to an academic question. 2.~A full solution for the question --- this is your primary grading reference. 3.~A grading rubric for the question --- use this as supplementary
scoring criteria alongside the model solution.

Read all documents carefully, including any diagrams or figures.

Grade the student's answer by comparing it directly against the full solution provided. Identify which key ideas, steps, or results the student captured correctly, partially, or missed entirely. Use the rubric to assign point
values to each component and ensure your scoring is consistent with the stated criteria.

Respond ONLY with a JSON object, no Markdown fences, no commentary outside the JSON. The JSON must have exactly two keys:

\noindent\{ \\
''score'': \textless{}integer between 0 and 100\textgreater{}, \\
\smallskip
''rationale'': ''\textless{}2-4 sentence explanation citing specific elements from the provided solution and grading rubric that the student addressed or missed in their answer \textgreater{}'' \\
\}
\end{quote}

\end{document}